\documentclass[10pt,twocolumn,letterpaper,table]{article}

\usepackage{cvpr}              % To produce the CAMERA-READY version
\usepackage{blindtext}
\usepackage[accsupp]{axessibility}
\usepackage{multirow}
\usepackage{booktabs} 
\usepackage{array}
\newcolumntype{Q}[1]{>{\centering\arraybackslash}p{#1}}
\usepackage{tikz} %for drawing figures
\usetikzlibrary{positioning}
\usepackage{tcolorbox}
\usepackage{caption}
\usepackage{graphicx}
\usepackage{subcaption}
\usepackage{float}  
\usepackage{amsmath}
\usepackage{ulem}
\usepackage{algorithm}
\usepackage{algpseudocode}
\usepackage{fix-cm}
\definecolor{codegreen}{rgb}{0,0.6,0}
\definecolor{codegray}{RGB}{192, 192, 192}
\definecolor{codepurple}{rgb}{0.58,0,0.82}
\definecolor{backcolour}{rgb}{0.95,0.95,0.92}
\definecolor{indaco}{RGB}{102, 178, 255}

\definecolor{mygreen}{rgb}{0,0.6,0}
\definecolor{mygray}{rgb}{0.5,0.5,0.5}
\definecolor{mymauve}{rgb}{0.58,0,0.82}
\definecolor{darkgray}{rgb}{.4,.4,.4}
\definecolor{navy}{HTML}{000080}
\definecolor{purple}{rgb}{0.65, 0.12, 0.82}
\definecolor{codepurple}{rgb}{0.58,0,0.82}
\definecolor{backcolour}{rgb}{0.95,0.95,0.92}
\definecolor{myblue}{RGB}{49, 104, 253} %search
\definecolor{myyellow}{RGB}{255, 205, 0}
\definecolor{mycyan}{RGB}{0, 222, 255}
\definecolor{mypurple}{RGB}{102, 0, 204}
\definecolor{mypink}{RGB}{255, 71, 250}
\definecolor{mywhite}{RGB}{255, 255, 255}
\definecolor{mygood}{rgb}{0.564705882352941,0.933333333333333,0.564705882352941}
\definecolor{mybad}{rgb}{0.980392156862745,0.501960784313725,0.447058823529412}
\newcommand{\argmin}{\arg\!\min}
\definecolor{cvprblue}{rgb}{0.21,0.49,0.74}
\usepackage[pagebackref,breaklinks,colorlinks,citecolor=cvprblue]{hyperref}

\def\paperID{*****} % *** Enter the Paper ID here
\def\confName{CVPR}
\def\confYear{2024}

\title{GRASP-GCN: Graph-Shape Prioritization for Neural Architecture Search under Distribution Shifts}

\author{Sofia Casarin$^1$,
Oswald Lanz$^1$ 
Sergio Escalera$^{2,3}$\\
$^1$Free University of Bozen-Bolzano, Bolzano, Italy \\
$^2$Computer Vision Center, Barcelona, Spain \\
$^3$Universitat de Barcelona, Barcelona, Spain \\
{\tt\small scasarin@unibz.it,  lanz@inf.unibz.it, sergio@maia.ub.es}
\and
}
\begin{document}
\maketitle
\begin{abstract}
Neural Architecture Search (NAS) methods have shown to output networks that largely outperform human-designed networks. However, conventional NAS methods have mostly tackled the single dataset scenario, incuring in a large computational cost as the procedure has to be run from scratch for every new dataset.
% --------- what we propose ---------
In this work, we focus on predictor-based algorithms and propose a simple and efficient way of improving their prediction performance when dealing with data distribution shifts. We exploit the Kronecker-product on the randomly wired search-space and create a small NAS benchmark composed of networks trained over four different datasets. To improve the generalization abilities, we propose GRASP-GCN, a ranking Graph Convolutional Network that takes as additional input the shape of the layers of the neural networks. GRASP-GCN is trained with the not-at-convergence accuracies, and 
% --------- what we achieve ------------
improves the state-of-the-art of 3.3 \% for Cifar-10 and increasing moreover the generalization abilities under data distribution shift.
\end{abstract}\vspace{-0.5cm}    
\section{Introduction}
%  What is NAS, what its impact and usefulness %
Neural Architecture Search (NAS) has drawn large research attention due to its efficacy in automatically optimizing the architecture of Deep Neural Networks (DNNs), replacing the error-prone manual design which demands high expertise. As the NAS process can be very expensive many methods were proposed to save time or computation, following two main directions: i) reducing the time required to evaluate each searched architecture proposing a weight sharing mechanism (\cite{cai2020onceforall},~\cite{bender18a},~\cite{pham2018efficient},~\cite{darts},~\cite{xie2020snas}), ii) using sample efficient algorithms so that only few architectures are evaluated (\cite{zoph2018},~\cite{zela2018},~\cite{klein2016},~\cite{real2019}). Proxy task performance and Predictor-based algorithms follow the second approach. They estimate the performance of the DNN either as an approximation or a prediction based on lower fidelities, such as i) shorter training (\cite{zoph2018},~\cite{zela2018}), ii) training on a subset of the data (~\cite{klein2016}), iii) on lower-resolution images (~\cite{chrabaszcz}), or iv) with less filters per layer and less cells (~\cite{zoph2018},~\cite{real2019}).While these approximations reduce the computational cost, they also introduce a bias in the estimate as performance will typically be underestimated.
%
% -------  predictor BASED algorithm and what's current SOTA ---------
Predictor-based algorithms follow the second approach, and train a proxy model that can infer the validation accuracy of DNNs directly from their network structure. During optimization, the proxy can be used to narrow down the number of architectures for which the true validation accuracy must be computed, which makes predictor-based algorithms sample efficient. Predictor-based algorithms have been proposed by~\cite{wen2019neural};~\cite{ning2020generic} and~\cite{BRP_NAS}.
%--------- What's the problem of NAS in general - comparison problems, of also predictor based algo and NAS ? dummy fact that procedure has to be run everytime for a different dataset, and  difficult reproducibility as computational intensive -- that's why focus on predictor based algorithms \\
Despite the success of these kinds of approaches, only few methods (\cite{lee2021rapid},~\cite{huang2022archgraph}) tackle the problem of sharing or re-using the predictor knowledge on different datasets. Most conventional NAS methods are indeed task-specific, requiring repeatedly training the model from scratch for each new dataset.
%----------- why we create our new dataset ------------
Moreover, existing NAS benchmarks either i) provide architectures trained on a single dataset
%
%\textbf{Chris Ying, Aaron Klein, Esteban Real, Eric Christiansen, Kevin Murphy, and Frank Hutter. 2019. NAS-Bench-101: Towards Reproducible Neural Architecture Search.02 2019}
(\cite{NASBENCH101}), ii) define a benchmark across different tasks
but not datasets (\cite{TransNAS}), iii) or do not provide the full training-log of architectures and define not-unique networks in their search space
%\textbf{(Xuanyi Dong, Lu Liu, Katarzyna Musial, and Bogdan Gabrys. 2020. NATS-Bench: Benchmarking NAS algorithms for Architecture Topology and Size. (08 2020).),}
(~\cite{NATSBENCH}). This
limit the possible studies that can be done on the datasets to correctly interpret possible predictor results. 
% ----------------------- what we propose ------------------
In this paper, we restrict the problem to predictor-based algorithms, which given a NAS benchmarks are extremely fast to train ($\sim$10 min on GeForce GTX 1080). We aim at answering this question: \textit{can we re-use knowledge that a predictor has learned on one dataset and transfer it to get a more sample-efficient algorithm on another dataset?} We study the impact of distribution shifts on the predictor performance, by analysing the ranking of the architectures trained on commonly used datasets for the task of image classification, and propose simple yet effective solutions to address the shift. Specifically, we design a randomly wired search space, that quaintly exploits the Kronecker product to impose a Resnet-like structure and create a dataset of 2000 architectures trained on Cifar-10, Cifar-100, Tiny-ImageNet and Fashion-MNIST datasets.
We study the generalization abilities of the predictor when directly used on a new dataset without fine-tuning and propose to integrate the so called \textit{vertex shapes} - the shapes each layer has given a different input size, and to adopt early stopping - training the predictor with not-converged accuracies. Our study shows which are the limitations of predictor based algorithms, and our simple approach improves of \textbf{ 3.7\%} and \textbf{9.5\%} (without and with distribution shift) with respect to the na\"ive approach . To summarize, our contributions are threefold:\begin{itemize}
    
    \item We propose a new way of defining search spaces, that exploit the generality of randomly wired spaces but samples neural network efficiently through the Kronecker product and a criterion based on a desired skeleton, to obtain specific categories of neural networks.
    \item We analyse the generalization capabilities of na\"ive predictor based algorithms on two different scenarios, involving different latent data but the same observed data and different latent data with different observed data.  
    \item We propose GRASP-GCN, which integrates the shapes of the layers of the neural networks as input to predictor, and trains with the accuracies of non-specialized neural networks. 
\end{itemize}
\begin{figure*}
    \centering
    \includegraphics[width=14.5cm,height=4.1cm]{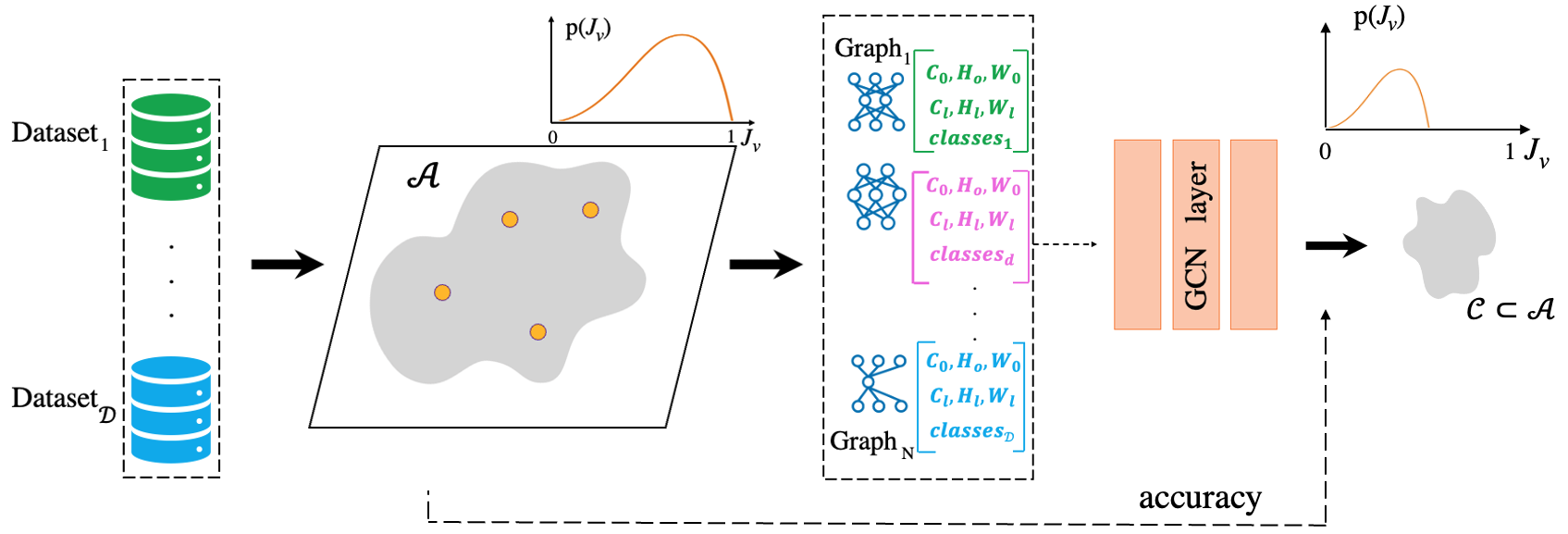}
    \caption{Architectures are sampled from the search space and trained over 4 datasets. The structure of DNNs is given as input with the shapes of the layers to a ranking GCN, which given the accuracy learns to rank DNNs so that the search space is narrowed down.}
    \label{fig:method}\vspace{-0.5cm}
\end{figure*}

\section{Related Works}\label{related_works}
Different techniques were proposed to mitigate the large computation burden of NAS. These approaches primarily target the acceleration of either the evaluation or search modules within the NAS framework. The former accelerates the evaluation of each DNN, the latter increases the sample efficiency so that fewer architectures need to be evaluated for discovering a good network. Our work falls under the second category, as the predictor can be utilized to sample architectures that most likely perform well on a given task.

\subsection{Single dataset predictor-based algorithms}
Many predictor-based methods, that set the baseline for following works, focus on a single dataset.
%In \textbf{cita 19 tesi} the authors train
\cite{wen2019neural} train a regressor model on a small built dataset and select the top-K predicted architectures to train them from scratch. The proposed approach leads to a more than 20$\times$ sample efficient algorithm with respect to standard used Evolution ones.
\cite{npenas2023} used graph neural network-based accuracy predictors and an iterative approach to estimate the accuracy of models. 
%udziak et al. propose in \textbf{cita [3] tesi} 
~\cite{wen2019neural} propose a Graph-based neural Architecture Encoding Scheme, \textit{i.e.} GATES, to improve the generalization abilities of performance predictors by modeling the information flow of the actual data processing of the architecture as the attributes of the input nodes. Neural Architecture Optimization, shortly NAO,~\cite{luo2019neural}) reframes the NAS problem as a continuous optimization problem. Through the use of a predictor that takes as input the continuous encoded representation of a neural network, NAO performs gradient based optimization in the continuous space to find the embedding of a new architecture with potentially better accuracy. 
%Despite the achievements in terms of efficiency and accuracy, none of these methods focus on multiple datasets analysing the actual transferability of predictors knowledge, and conduct an analysis on the performance of predictors when trained with data coming from architectures not trained until convergence. 
~\cite{BRP_NAS} propose an efficient hardware-aware NAS method enabled by an accurate performance predictor based on Graph Convolutional Network (GCN). The authors show that the sample efficiency of predictor based NAS can be improved by considering binary relations of models and an iterative data selection strategy. Similarly to BRP-NAS, we employ a binary ranking GCN, but we extend the focus on multiple datasets and employ as labels the validation accuracy of non-specialized networks to improve the generalization abilities. 
\subsection{Transferable predictor-based algorithms}
Among existing methods, relevant approaches to ours tackling the generalization problem across multiple dataset are MetaD2A by~\cite{lee2021rapid}, and Arch-Graph by~\cite{huang2022archgraph}. 
MetaD2A stochastically generates graphs from a dataset via a cross-modal latent space that is learned via amortized meta-learning. From the encoding of the dataset, obtained through a permutation invariant encoder set, a graph is decoded. A meta-predictor is then used to estimate and select the best architecture for a given dataset. Instead of using an encoder set, we propose a much simpler and more general solution that does not limit the approach to image-dataset due to the encoder-set, lacking the possibility to adapt it to video. We provide as additional input to the GCN the ``vertex shapes", which are strictly related to the shapes of the data. In Arch-Graph the generalization problem is addressed from the point of view of task generalization, rather than dataset generalization. The method predicts task-specific optimal architectures with respect to given task embeddings, by leveraging correlations across multiple tasks through their embeddings as a part of the predictor’s input for fast adaptation. Despite being sample efficient across many tasks, the method requires predictor-tuning on the new task/dataset. 
With respect to previous approach, our work tackles the problem from the point of view of the distribution shift in the dataset, proposing a simpler yet more general method that takes into consideration the specialization (or overfitting) of neural networks over datasets, the shape characteristics of the data and their effect on networks, and that does not require fine-tuning the predictor.
\section{Methods}
Our goal is to obtain a predictor that generalizes well across different datasets without need to re-train or finetune. To this end, we sample architectures from our search-space and train them over 4 image classification datasets (\ref{subsec:search-space}) and propose GRASP-GCN (\ref{subsec:gcn}), which trains a ranking predictor with an additional input consisting in the vertex shapes.
\subsection{Search-space definition}\label{subsec:search-space}
NAS is formalized as a bi-level optimization problem:
\begin{equation}
\label{eq:bi-level}
\begin{aligned}
\mathcal{A^*}  =  \argmin{_\mathcal{A}} J{_v}  ( \mathcal{A}, \textbf{w}^{*})  \\
\quad \textrm{s.t.} \quad  \textbf{w}^{*} = \argmin{_\textbf{w}} J_t (\mathcal{A}, \textbf{w}),
\end{aligned}
\end{equation}
where \( \mathcal{A} \) describes the architecture, \( \textbf{w}\) are the trainable weights of the considered DNN, and $J_v$ and $J_t$ are the validation and training loss, respectively. In our work we define \( \mathcal{A} = ( \mathbf{A}, \mathbf{X}) \), with \( \mathbf{A}^{N\times N} \) and \( \mathbf{X}^{N\times D} \) encoding the connections in the graph and the types of layers, respectively. The dimensions of the matrices are related to the N number of nodes (layers) and the D number of input features, \textit{i.e.} the number of layer types allowed.
% ----------------------------------------------------
We sample $\mathcal{A}$ from our search space composed of all feed-forward convolutional networks, belonging to the randomly-wired search space~\cite{randomly_wired} and exploit the Kronecker product, a trick that allows us balancing flexibility and efficiency, avoiding the sampling of big random matrices.
We generate resnet-like architectures miming evolutionary sampling. Two random matrices \(\mathbf{R}_1^{4\times4}\) and \(\mathbf{R}_2^{4\times4}\) are indeed sampled from the search space and multiplied with two so-called skeleton matrices \(\mathbf{K}_1^{4\times4}\) and \(\mathbf{K}_2^{4\times4} \):
\begin{equation}
\resizebox{0.6\linewidth}{!}{%
$
\begin{aligned}
\mathbf{A} = \quad &
\underbrace{
\begin{bmatrix}
1 & 0 & 0 & 0 \\
0 & 1 & 0 & 0 \\
0 & 0 & 1 & 0 \\
0 & 0 & 0  & 1 \\
\end{bmatrix}}_{{\color{gray}\text{K$_1$}}}
\otimes
\begin{bmatrix}
 & & & & \\
 & & \multicolumn{2}{c}{\multirow{2}{*}{\text{\(R_1\)}}} & & \\
 & & & & \\
  & & & & \\
\end{bmatrix}
 \\
+ \quad & 
\underbrace{
\begin{bmatrix}
0 & 1 & 0 & 0 \\
0 & 0 & 1 & 0 \\
0 & 0 & 0 & 1 \\
0 & 0 & 0  & 0 \\
\end{bmatrix}}_{{\color{gray}\text{K$_2$}}}
\otimes
\begin{bmatrix}
 & & & & \\
 & & \multicolumn{2}{c}{\multirow{2}{*}{\text{\(R_2\)}}} & & \\
 & & & & \\
  & & & & \\
\end{bmatrix}
\\
=  \quad &
\begin{bmatrix}
    R_1 &R_2 & \multicolumn{1}{c}{} \\
    & R_1 & R_2 & \\
    & & R_1 & R_2 \\
    \multicolumn{3}{c}{}  & R_1 \\
\end{bmatrix}
\end{aligned}$%
}
\label{eq:kron_detail}
\end{equation}

In Eq.~\ref{eq:kron_detail}, \(\mathbf{K_1}\) which is multiplied by \(\mathbf{R_1}\) defines the feed-forward structure, while \(\mathbf{K_2}\), with the off-diagonal values, defines the shortcuts. The \textit{input} and the \textit{output} layers are finally added to the generated matrix \( \mathbf{A}^{16\times 16} \), leading to a maximum dimension of 18$\times$18. 
As shown, the Kronecker product $\otimes$ allows repeating \(\mathbf{R}_1\) and \(\mathbf{R}_2\) structures, limiting the randomly wired search space in a meaningful way.  Moreover, it generates easily scalable networks, a key advantage as proven by~\cite{zoph2018}, by stacking multiple blocks. This design choice was intentional to focus on analyzing the true impact of the dataset itself. Moreover, it does not represent a limitation as the only requirement to properly train a GCN predictor is to have both very-well and very badly-performing DNNs. Table~\ref{tab:layers} summarized our candidate layers which constitute our $ \mathbf{X}^{18\times 9} $.

\begin{table*}
\centering\fontsize{9}{11}\selectfont
\begin{tabular}{ |c|c|c|c|c|c|c|c| } 
\hline
Layer & stem conv3$\times$3 & conv3$\times$3 & conv3$\times$3$\times$d & conv3$\times$3$\times$h & conv3$\times$3s2 & conv3$\times$3s2$\times$d & conv3$\times$3s2$\times$h \\ \hline 
Channels& 64 & \textbf{Same} & \textbf{Doubles} & \textbf{Halves} & \textbf{Same}  & \textbf{Doubles}  & \textbf{Halves} \\ \hline
Stride & - & - & - & - & 2 & 2 & 2 \\ \hline
\end{tabular}
\caption{Layers in the search space. Same / Double / Halves refer to the channels of the parent nodes.}
\label{tab:layers}\vspace{-0.5cm}
\end{table*}

\begin{comment}
    
\bigskip

\begin{tabular}{ |>{\centering\arraybackslash}p{2.6cm}|>{\centering\arraybackslash}p{2.6cm}|>{\centering\arraybackslash}p{2.6cm}|>{\centering\arraybackslash}p{2.6cm}|>{\centering\arraybackslash}p{2.6cm}|>{\centering\arraybackslash}p{2.5cm}| } 
\hline
Layer type & conv3$\times$3s2 & conv3$\times$3s2$\times$d & conv3$\times$3s2$\times$h & output layer \\ \hline 
Channels & \textbf{Same}  & \textbf{Doubles}  & \textbf{Halves} & - \\ \hline
Stride & 2 & 2 & 2 & - \\ \hline
\end{tabular}
\caption{Layers in the search space. Same / Double / Halves refer to the channels of the parent nodes.}
\label{tab:layers}
\end{comment}

\subsection{Ranking GCN}\label{subsec:gcn}
Graph Convolutional Networks are DNN architectures that extract multi-scaled localized spatial features to extract highly expressive representations of graphs, dealing with the difficulty of ``localized convolution" filters in non-Euclidean domains. GCNs performs a convolution looking for essential vertices and edges with the goal of learning the features of the graph. It takes as input: (i) a feature description X$_i$ for every node \textit{i} summarized in a feature matrix \( \mathbf{X}^{N\times D} \) where N is the number of nodes, D the number of input features; (ii) a representative description of the graph structure summarized in the adjacency matrix \( \mathbf{A}^{N\times N} \). For the classification task, the GCN produces a node-level output \( \mathbf{H}^{N\times F} \), where F is the number of output features per node. The GCN outputs $h(\mathbf{A,X)}$ which is the concatenation of each \( \mathbf{H}^l \) layer mapping done, as described in Eq.~\ref{eq:concat}:
\begin{equation}
\label{eq:concat}
\begin{aligned}
h(\mathbf{A}, \mathbf{X}) = (H^L  \circ H^{L-1}  \circ \dots \circ H^l \dots \circ H^1 ) (\mathbf{A}, \mathbf{X})
\end{aligned}
\end{equation}

where \(l\) = 1, 2, .. L, is the number of layers in the GCN, and  
\(\mathbf{H}^l\) is given by the propagation rule (eq. \ref{prop_rule}) defined by in \cite{kipf2017semisupervised}:

\begin{equation}\label{prop_rule}
    H^{(l+1)} = \sigma(\tilde{D}^{-\frac{1}{2}}\tilde{A}\tilde{D}^{-\frac{1}{2}}H^{(l)}W^{(l)})
\end{equation}
with \( \tilde{A} = A + I \), ($I$ identity matrix), \( \tilde{D}_{ii} = \sum_{j}\tilde{A}_{ij} \) is the degree matrix, $W^{(l)}$ is the layer-specific trainable weight matrix, $\sigma$ is the ReLU activation function, $H^{(l)} \in \mathbb{R}_{N \times D}$ is the matrix of activations in the $l^{(th)}$ layer, and $H^{(l+1)}$ is the output to the next layer. Graph-level outputs can then be modeled by introducing some form of pooling operation. 
In our work we exploit a GCN with 3 layers and a classification head, and following BRP-NAS we exploit a Ranking GCN, which learns a binary relation that focuses on the prediction of ranking. Indeed, as previously observed by~\cite{BRP_NAS} \textit{i)} accuracy prediction is not necessarily required to produce faithful estimates (in the absolute sense) as long as the predicted accuracy preserves the ranking of the models; \textit{ii)} any antisymmetric, transitive and convex binary relation produces a linear ordering of its domain, implying that NAS could be solved by learning binary relations, where $O(n^2)$ training samples can be used from \emph{n} measurements.
Given the architecture's search space, a ranking network predicts \textit{how likely} any network in the search space reaches a higher accuracy than the current best.
\begin{figure*}
    \centering
    \includegraphics[width=14.5cm,height=2.7cm]{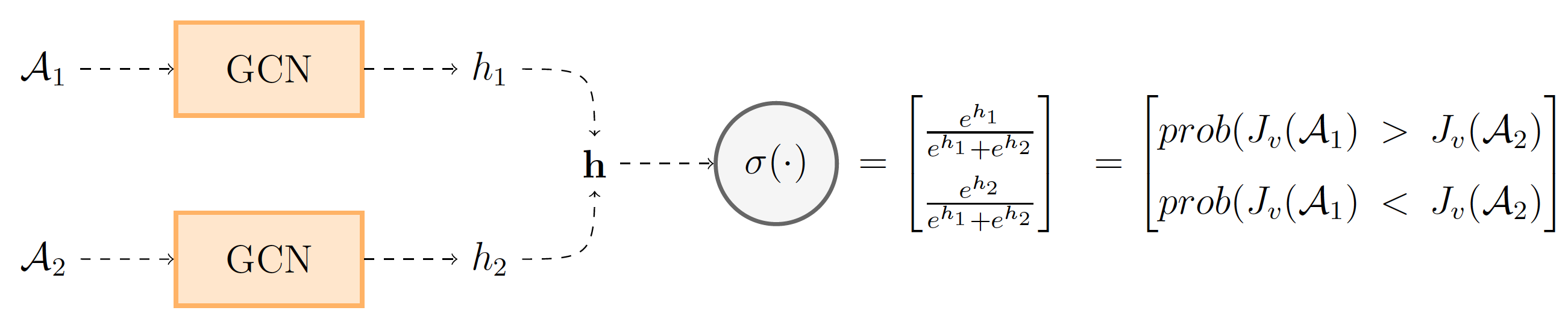}
    \caption{Working principles of a GCN used as a ranking network}
    \label{fig:method_GCN}
\end{figure*}
Fig. \ref{fig:method_GCN} shows how a GCN can be used as a ranking network: two architectures $\mathcal{A}_1$ and $\mathcal{A}_2$ are fed to the GCN. The GCN outputs two graph encodings h$_1$ and h$_2$, which are then concatenated into the vector $\mathbf{h}$. A softmax function $\sigma()$ is applied, obtaining the ranking probabilities which are then compared with the target \(\mathbf{t}\). Given for example the tuple (\( acc_{i,\mathcal{A}_j}, acc_{i,\mathcal{A}_k}) \) where \(acc_{i,\mathcal{A}_j}\) is the accuracy of architecture \( \mathcal{A}_j\) and \(acc_{i,\mathcal{A}_k} \) is the accuracy of \(\mathcal{A}_k\), both over dataset$_i$, if \( acc_{i,\mathcal{A}_j} > acc_{i,\mathcal{A}_k} \), then 
the target vector takes as values $\mathbf{t} = [1 \quad 0]$. Our goal is to maximize the log probability that $\mathcal{A}_j$ is better than $\mathcal{A}_k$, and therefore the predictor is trained with the loss in Eq.~\ref{eq:rank_target}:
\begin{equation}\label{eq:rank_target}
    J = - \mathbf{t}^\intercal \cdot ln(\sigma(\mathbf{y})). 
\end{equation}

As shown in Fig.~\ref{fig:method_GCN}, $\mathcal{A}_{j,k}$, which gives the information about the structure of the graphs, and the feature vector $\mathcal{X}$, which defines the layers in the DNN architecture. Therefore, given a DNN architecture that is characterized by a bottleneck, the predictor won’t capture the implications of a severe reduction of the feature maps. Clearly, the same DNN architecture would have very different performance depending on the input shape and the predictor could not see the difference if given with $\mathcal{A}$ and $\mathcal{X}$ only. We therefore add the dimensions characterizing each DNN layer. This info shortly called ``\textit{vertex shapes}'' and is concatenated to $\mathcal{X}$ that has now dimensions 18$\times$12. We normalize the vertex shapes with respect to the maximum dimensions in our dataset and encode them as float numbers. We did not choose one hot encoding as it implicitly looses any ordering and distance knowledge, \textit{i.e.} we don't know if shape (3,32,32) is closer to (3,64,64) or to (4096,3,3).

\section{Implementation details}
\label{app:impl_det}
Our hyper-parameters are summarized in Tab.~\ref{tab:hyper-resnet18} which displays those used for training the architectures sampled from our search-space and in Tab~\ref{tab:hyper-cifar10} for training our GRASP-GCN. All hyper-parameters were optimized using Optuna framework\footnote{https://optuna.readthedocs.io/en/stable/}. 
We moreover provide the pseudo-code for the creation of our NAS dataset (\cref{algo:dataset}), and for the functioning of predictor-based algorithms within the NAS framework (\cref{algo:gcn_nas}).\vspace{-0.3cm}

\begin{table*}[t]
\centering
\begin{tabular}{ |c|c|c|c|c|c| } 
\hline
Dataset & Learning rate & Weight decay & Drop lr & Optimizer & Batch-size \\
\hline
F-MNIST & 0.100 & 0.0002 & of 10 at epoch 40, 80 & SGD & 128 \\
\hline
C10 & 0.097 & 0.0006 & of 10 at epoch 40, 80 & SGD & 128  \\
\hline
C100 & 0.065 & 0.0015 & of 5 at epoch 40, 80, 100  & SGD  & 128\\
\hline
Tiny & 0.012 & 0.0011 & of 10 at epoch 30, 60, 90 & SGD & 64\\
\hline
\end{tabular}
\caption{List of hyper-parameters derived from Optuna optimization for ResNet-18. The columns show the learning rate (lr), the weight decay (wd), the drop of the learning rate (drop lr), the optimizer, and the size of the batches.}
\label{tab:hyper-resnet18}\vspace{-0.4cm}
\end{table*}

\begin{table}
\centering\fontsize{9.5}{11}\selectfont
\begin{tabular}{ |c|c|c|c| } 
\hline
Units per layer & Lr & Weight decay & Optimizer \\
\hline
265 & 0.019041 & 0.001126 & Adagrad \\
\hline
\end{tabular}
\caption{List of hyper-parameters derived from Optuna optimization for GRASP\_GCN.}
\label{tab:hyper-cifar10}\vspace{-0.3cm}
\end{table}

\begin{algorithm}
	\caption{Dataset creation} 
	\begin{algorithmic}[1]
	   \If {$Cifar10$}
	        \For {$iteration=1,\ldots, 2000$}
            		    
		        \State sample $\mathcal{A} =  \mathbf{A}, \mathbf{X}$
		        \State train $\mathcal{A}$
		        \State save values in hash($\mathcal{A}$)
		    \EndFor
	   \EndIf
        \If {FashionMNIST or Cifar100 or TinyImageNET}
            \For {$iteration=1,\ldots, 2000$}
                \State random sample $\mathcal{A}$ from Cifar10 dataframe
                \State new directory = hash($\mathcal{A}$)
                \While {new directory exist}
                    \State random sample $\mathcal{A}$ from Cifar10 dataframe
                    \State new directory = hash($\mathcal{A}$)
                \EndWhile
                \State train $\mathcal{A}$
                \State save values in new directory
            \EndFor
        \EndIf
	\end{algorithmic}
 \label{algo:dataset}
\end{algorithm} \vspace{-0.5cm}

\begin{algorithm}
	\caption{Predictor-based neural architecture search} 
	\begin{algorithmic}[1]
	\State $\mathcal{S}$ = Architecture search space
	\State \(f\mathcal {(A, \theta)} \) : $\mathcal{A}\rightarrow \mathbb{R}$: Predictor that outputs the predicted performance given the architecture
	\State $N^{(k)}$: Number of architectures to sample in the k-th iteration
	\State k = 1
	\State $\Tilde{\mathcal{S}} = \varnothing$
	\While{$k \leq$ MAX\_ITER}
	    \State Sample a subset of architectures $\mathcal{C}^{(k)}$ = $\{\mathcal{A}_j^{(k)}\}_{j=1, \dots, N^{(k)}}$ from $\mathcal{S}$ utilizing \(f\mathcal {(A, \theta)} \)
	    \State Evaluate architectures in $\mathcal{S}^{(k)}$ , get $\Tilde{\mathcal{C}}^{(k)}$ = $\{\mathcal{A}_j^{(k)}, y_j^{(k)} \}_{j=1, \dots, N^{(k)}}$ ($y$ is the performance)
	    \State $\mathcal{S} = \mathcal{S} - \mathcal{C}$
	    \State $\Tilde{\mathcal{S}} = \Tilde{\mathcal{S}} \cup \Tilde{\mathcal{C}}$
	    \State Optimizing \(f\mathcal {(A, \theta)} \) using the ground-truth architecture evaluation data $\Tilde{\mathcal{S}} $ 
	\EndWhile
	\State Output $\mathcal{A}_{j^{*}}$ $\in$ $\Tilde{\mathcal{S}}$ with best corresponding $y{^*}$; Or, $\mathcal{A}^{*}$ = $argmax_{\mathcal{\Tilde{S}}}$\( f \mathcal {(A, \theta)} \)
	\end{algorithmic} \label{algo:gcn_nas}
\end{algorithm} 

\section{Experiments }
% ----------------------------------- groun truth ----------------------------------
In this section, we introduce the used ranking measures (\cref{app:ran_meas}), we investigate the statistics of our NAS dataset composed of (architectures, accuracy) pairs to provide ground-truth values for the predictor (Sec.~\ref{subsec:dataset}). We show the performance of our predictor under distribution shift (Sec.~\ref{subsec:grasp_gcn}) and further validate the usage of the vertex shapes as additional input and of not converged accuracies through ablations (Sec.~\ref{subsec:ablation}). %Our performance are reported with three different measures: validation accuracy, Precision@10, and Normalized Discounted Cumulative Gain (NDCG). 
\vspace{-0.3cm}

\section{Ranking measures}
\label{app:ran_meas}

Several measures are commonly used to assess the quality of a ranking.
We focus on: 1) NDCG@k, 2) Precision@k and 3) Kendall's $\tau$.  
\paragraph{NDCG}
DCG (Discounted Cumulative Gain) is founded on the idea that when assessing search results, highly relevant documents ranked lower should receive a larger penalty than less relevant documents wrongly ranked. This is because the graded relevance value decreases logarithmically as the position of the result worsens. This measure applies very well to predictor-based algorithms since our focus is on allowing the predictor to find the best-forming architectures in such a way that they are placed on the top of the list, while we don't care about how bad networks are ranked.
Eq.~\ref{eq:DCG} exactly shows how highly relevant objects that appear lower in the ranked list are penalized by reducing the graded relevance value logarithmically proportional to the position of the result:

\begin{equation}
\label{eq:DCG}
\begin{aligned}
DCG_k = \sum_{i=1}^{k} \frac{2^{rel_i} - 1}{log_2(i+1)} .
\end{aligned}
\end{equation}\

Starting from~\ref{eq:DCG}, the NDCG can be easily obtained normalizing w.r.t the Ideal Discounted Cumulative Gain (IDCG), as shown in Eq. \ref{eq:NDCG}:

\begin{equation}
\label{eq:NDCG}
\begin{aligned}
NDCG_k & = \frac{DCG_k}{IDCG_k} \in [0,1],  \\
IDCG_k & =  \sum_{i=1}^{|REL_k|} \frac{2^{rel_i} - 1}{log_2(i+1)}
\end{aligned}
\end{equation}

where $rel_i$ is the relevance value assigned to each object and $REL_k$  is the list of relevant objects ordered by their relevance up to position $k$. In a perfect ranking algorithm, the ${\displaystyle DCG_{p}}$ will be the same as the ${\displaystyle IDCG_{p}}$ producing an NDCG of 1.
We used the NDCG in two different variants: the NDCG@2092 was used to get a big picture of the general behavior all architectures have, while the NDCG@10 was used to focus on the ranking quality of the top-10 performing architectures, which are those we are interested in.

\paragraph{Precision@k} It is defined as the proportion of recommended items in the top-k set that are relevant. It is mathematically defined as:
\begin{equation}\label{eq:precision}
    Precision@k = \frac{\# \text{of items @k that are relevant}} {k} \in [0,1],
\end{equation}
where again, as it can be observed, the concept of relevance is involved. 

\begin{table}[H]
    \centering
    \fontsize{7.5}{11}\selectfont
    \begin{tabular}{ |c|c|c|c|c| } 
        \hline
        & F-mnist & C10 & C100 & Tiny \\
        \hline 
        F-mnist &\cellcolor{mygreen!70} 0 & \cellcolor{mygreen!40}0.0273 &\cellcolor{mygreen!25}0.0461 &\cellcolor{mygreen!25} 0.0516 \\ \hline
        C10  & \cellcolor{mygood!20} 0.1781 & \cellcolor{mygreen!70}0 &\cellcolor{mygood!40} 0.1032 &\cellcolor{mygood!30} 0.1293 \\  \hline
        C100 & \cellcolor{red!40}0.4559 &\cellcolor{myyellow!20} 0.2653 &\cellcolor{mygreen!70} 0 &\cellcolor{mygood!20} 0.1743 \\  \hline
        Tiny  & \cellcolor{red!50}0.5157 & \cellcolor{orange!20}0.3456 & \cellcolor{mygood!20} 0.1817 &\cellcolor{mygreen!70} 0  \\
        \hline
    \end{tabular}
    \caption{1-NDCG@2092. The columns display the training dataset, rows the validation dataset. Color bar: \includegraphics[height=0.2cm]{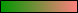} ({\color{mygreen!70}lowest} to {\color{red!50}highest}).}
    \label{tab:NDCG@2092}
\end{table}

\paragraph{Kendall's $\tau$} The Kendall rank correlation coefficient is used to measure the ordinal association between two measured quantities. 
It's range is in $[-1,1]$, with Kendall's $\tau$ = 0 that indicates absence of correlation. 
Let $(x_1, x_2, \dots x_n)$ and $(y_1, y_2, \dots y_n)$ be a set of observations of the joint variables \textit{X} and \textit{Y}, such that all the values of ($x_i$) and ($y_i$) are unique (ties are neglected for simplicity). Any pair of observations ($x_i, y_i$) and ($x_j, y_j$), where $i < j$ are said to be \emph{concordant} if the sort of order ($x_i, x_j$) and ($y_i, y_j$) agrees: that is, if either both $x_i > x_j$ and $y_i > y_j$ holds if both $x_i < x_j$ and $y_i < y_j$; otherwise, they are said to be \emph{discordant}. Eq. \ref{eq:kendall} defines, based on this, the coefficient:
\begin{equation}\label{eq:kendall}
   \tau = \frac{(\# \text{of concordant pairs}) - (\# \text{of discordant pairs})} {\binom{n}{2}},
\end{equation}

\subsection{Dataset}\label{subsec:dataset}

The four datasets were chosen based on the possibility to exploit a variety of conditions to investigate the transferability of the predictor, and by the desire to extend and be partially comparable with previous works that involve three datasets, as in NATS-Bench~\cite{NATSBENCH}. Two different scenarios were defined: i) transferability when different latent data is involved \textit{but} the observed data is \textbf{the same}, ii) transferability when different latent data is involved \textit{and} the observed data is \textbf{not the same}. Hence, we chose (\emph{a}) Cifar-10 as a baseline dataset, (\emph{b}) Cifar-100 as it is composed of Cifar-10 images, labeled differently, (\emph{c}) Fashion-MIST as it is composed of black and white images (different latent data), belonging to completely different categories with respect to Cifar-10, and (\emph{d}) Tiny-ImageNET to consider a more complex dataset with a larger number of classes. 2000 \textit{unique} architectures were sampled from the search space and trained over the aforementioned datasets.
\begin{figure}
\centering
    \begin{subfigure}{.3\textwidth} 
        \centering 
        \includegraphics[width=\textwidth]{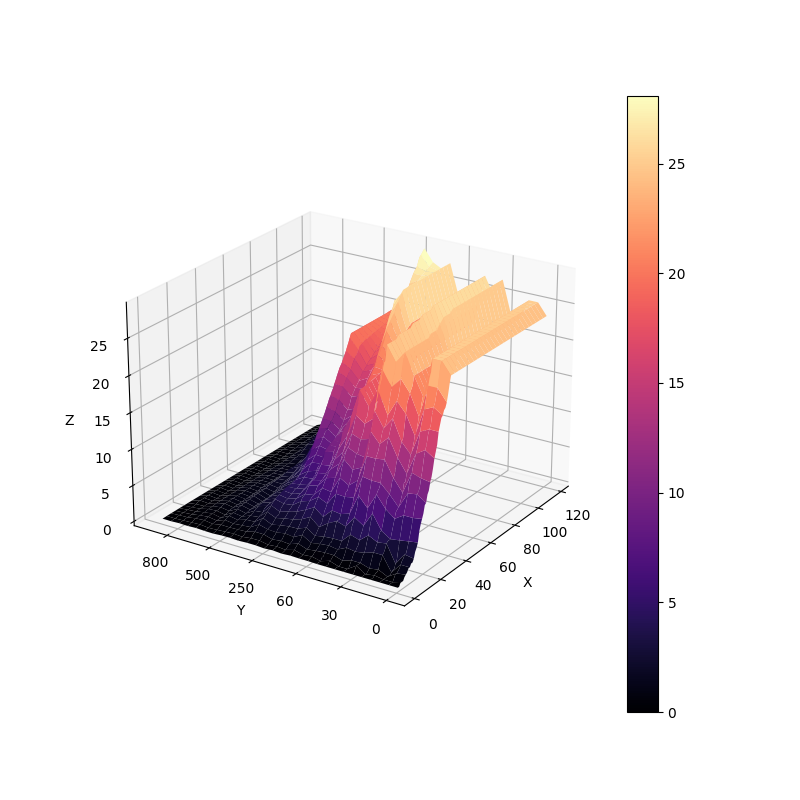}
        \caption{} 
        \label{fig:evo-rankingCUM}
    \end{subfigure} 
    \\
    \begin{subfigure}{.3\textwidth} 
        \centering 
        \includegraphics[width=\textwidth]{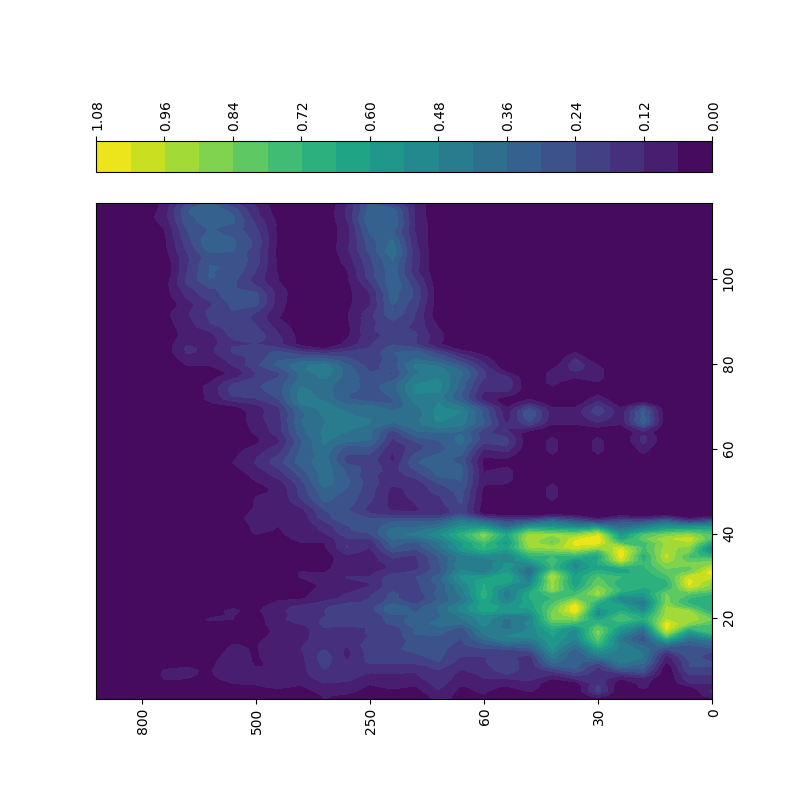}
        \caption{} 
        \label{fig:evo-rankingDEV}
    \end{subfigure} 
    \\
    \begin{subfigure}{.3\textwidth}  % Adjust the width as needed
        \centering
        \includegraphics[width=\textwidth]{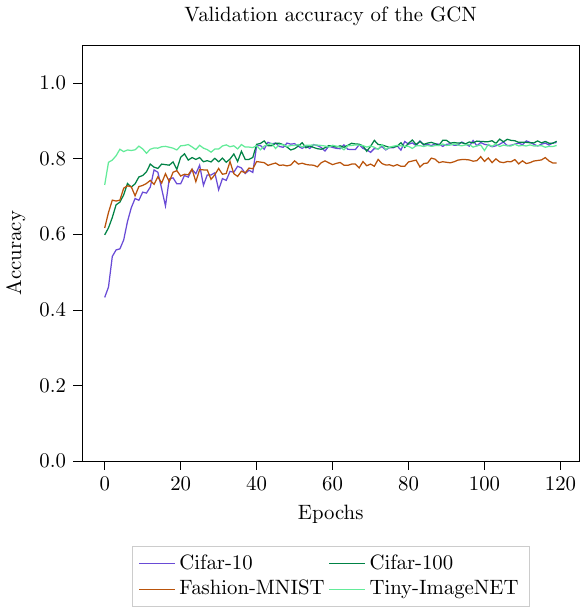}
        \caption{}
        \label{fig:early_stopping}
    \end{subfigure}
    \caption{Cifar-10 results. Ranking evolution during training with a cumulative (a) and derivative (b) plot. The x-axis shows the architectures in a descent order with respect to their accuracy. The y-axis carries the training epochs. The heatmap displays large numbr of rank changes (yellow) to no changes (blue). (c) Validation accuracy the GCN can obtain when trained with the previous rankings.} 
    \label{fig:cifar-10_results}\vspace{-0.5cm}
\end{figure}

\begin{figure}[htbp]
    \centering
    \includegraphics[width=5cm,height=3.2cm]{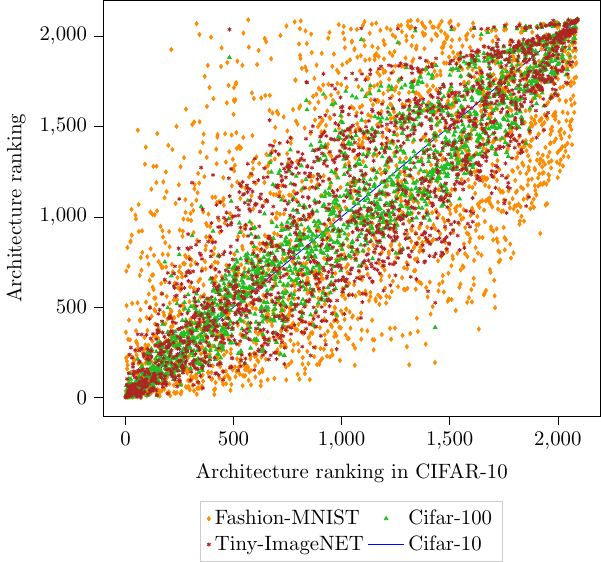}
    \caption{The ranking of each architecture on one of the four datasets, sorted by the ranking in Cifar-10. Correlation measured through Kendall's$\tau$.}
    \label{fig:NATS_plot}\vspace{-0.7cm}
\end{figure}
% ----------------------------------------- Statistics ----------------------------------------------
\begin{table*}[tb]
    \begin{minipage}[t]{0.5\textwidth}
        \centering
        \begin{subtable}[t]{\textwidth}
            \centering\fontsize{7.5}{11}\selectfont
            \begin{tabular}{ |c||c|c|c|c| } 
                \hline
                & F-mnist & C10 & C100 & Tiny\\
                \hline
                F-mnist & 85.1 $\pm$ 0.7  & 76.4 $\pm$ 0.5 & 68.2  $\pm$ 0.6 &68.4 $\pm$ 0.8 \\ \hline
                C10  &74.2 $\pm$ 0.6   & 87.9 $\pm$ 0.4 & 84.9 $\pm$ 0.4&   83.4 $\pm$ 0.7   \\  \hline
                C100   & 70.8 $\pm$ 0.6& 85.0 $\pm$ 0.1 & 87.0 $\pm$ 0.4  & 85.6 $\pm$ 0.3  \\ \hline
                Tiny & 72.3 $\pm$ 0.3 & 83.6 $\pm$ 0.9  & 85.3 $\pm$ 0.2  & 87.9 $\pm$ 0.2 \\
                \hline
            \end{tabular}
            \caption{Average validation accuracy (\%) of the predictor.}
            \label{tab:vs_acc}
        \end{subtable}
    \end{minipage}%
    \quad
    \begin{minipage}[t]{0.5\textwidth}
        \begin{subtable}[t]{\textwidth}
            \centering\fontsize{7.5}{11}\selectfont
            \begin{tabular}{ |c||c|c|c|c| } 
                \hline
                & F-mnist & C10 & C100 & Tiny \\
                \hline
                F-mnist&  100 $\pm$ 0 & 93 $\pm$ 2 & 94 $\pm$ 2 & 96 $\pm$ 1  \\ \hline
                C10  & 80 $\pm$ 4  & 99 $\pm$ 1 & 97 $\pm$ 2  &  95 $\pm$ 3 \\  \hline
                C100   & 36 $\pm$ 3 & 60 $\pm$ 2 &{65 $\pm$ 1} &{63 $\pm$ 1} \\ \hline
                Tiny  & 41 $\pm$ 2 &{61 $\pm$ 3} & {57 $\pm$ 3} & {69 $\pm$ 2}  \\
                \hline
            \end{tabular}
            \caption{Average Precision@10 ($\times$100) of the predictor.}
            \label{tab:vs_prec10}
        \end{subtable}
        \label{tab:vs}
    \end{minipage}
    \caption{Performance of our GRASP-GCN provided with input matrix \textbf{A}, features \textbf{X}, vertex shapes \textbf{V} and trained with the validation accuracy at epoch 40. For each element of the table, the mean and the std of four runs are given. The columns are the training datasets, the rows are the validation datasets.}
    \label{tab:combined_tables}\vspace{-0.5cm}
\end{table*}
Models trained over different input distributions are not guaranteed to perform in the same way.  Do they ``overfit" the dataset specializing during training? Can early stopping be applied to reduce the training time it takes to get the true validation accuracy of the architectures? 

\textbf{Distribution Shift} Tab.~\ref{tab:NDCG@2092}, to be read column-wise, highlights that the order induced by compelx datasets generalizes better for simpler datasets than vice-versa. This result is complemented by Fig.~\ref{fig:NATS_plot}, that displays the ranking stability of each architecture on one of the four datasets. The plot was obtained by sorting the architectures trained on Cifar-10 in descending order with respect to their accuracy on the dataset, and inducing the same ID-order on the architectures trained on different datasets. In this way, it is possible to observe what rank has been assigned to the architectures trained over different datasets, having as baseline Cifar-10 rank. We can observe that (i) Fashion-MNIST architectures cause a higher variability in terms of architectures rank ii) the plot start tighter, increase variance as we move towards bigger ranks, and gets tighter as we approach the ``worst'' architectures. We can deduce that, not only the worst architectures are such since the very beginning of training, (Fig.~\ref{fig:evo-rankingCUM},~\ref{fig:evo-rankingDEV}), but as it could be naturally expected, some architectures simply do not have enough capacity to solve any classification task and perform badly independently on the input distribution they are provided, further validating the early stopping method we propose in Sec.~\ref{subsec:grasp_gcn}.\\
\begin{figure}[tb]
 \begin{subfigure}{.23\textwidth} 
         \centering 
         \includegraphics[width=\textwidth]{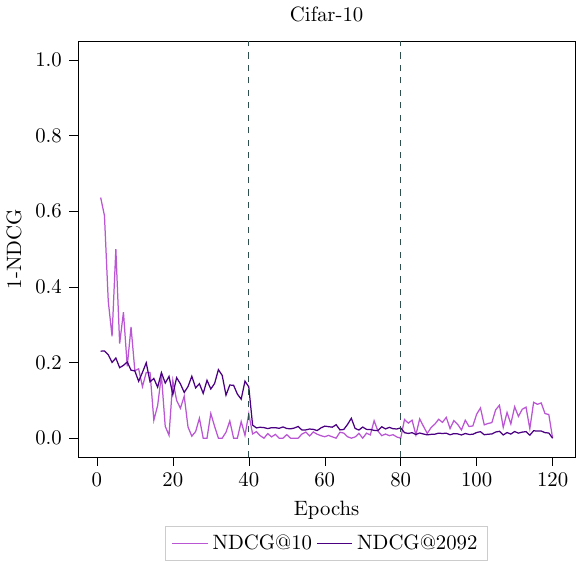}
         \caption{Cifar-10} 
         \label{subfig:cifar10_NDCG}
     \end{subfigure} \hfill
     \begin{subfigure}{.23\textwidth} 
         \centering 
         \includegraphics[width=\textwidth]{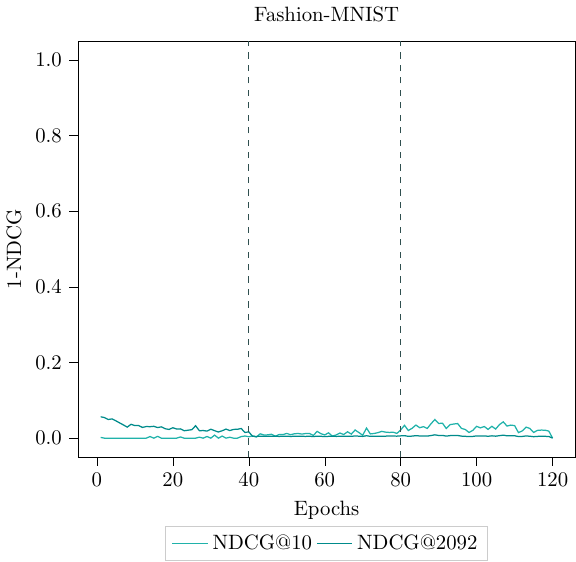}
         \caption{Fashion-MNIST} 
         \label{subfig:fashion_NDCG}
     \end{subfigure} 
     \hfill
     \begin{subfigure}{.23\textwidth} 
         \centering 
         \includegraphics[width=\textwidth]{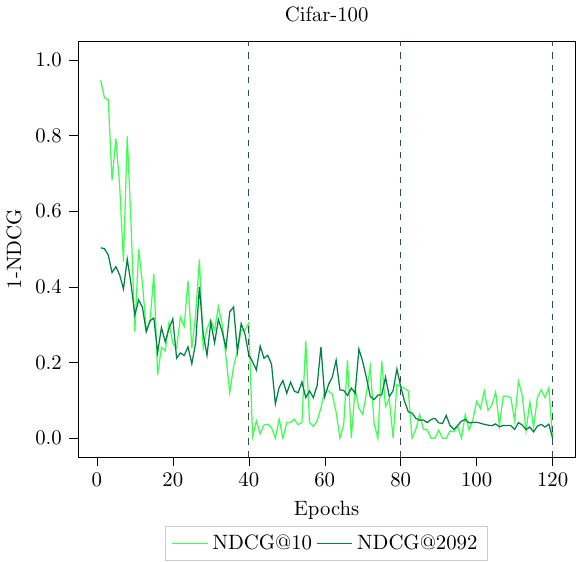}
         \caption{Cifar-100} 
         \label{subfig:cifar100_NDCG}
     \end{subfigure}% 
     \hfill
     \begin{subfigure}{.23\textwidth} 
         \centering 
         \includegraphics[width=\textwidth]{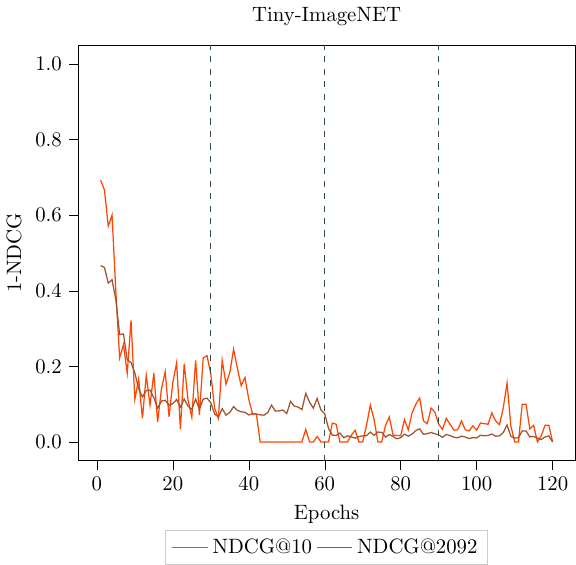}
         \caption{Tiny-ImageNET} 
         \label{subfig:tiny_NDCG}
     \end{subfigure} 
     \caption{1-NDCG plot showing the ranking correlation among the sorting the architectures have at epoch $i$ w.r.t the sorting of the architectures at epoch 120. The lower the better. Every figure displays the results for considering one dataset at a time. The dashed lines over each plot highlight the epoch where the learning rate is dropped. Every plot displays both the NDCG@10 (light color) and the NDCG@2092 (dark color).} 
     \label{fig:NDCGat10_2092}\vspace{-0.7cm}
\end{figure} 
 \begin{table*}
    \centering
    \begin{subtable}{\textwidth}
        \centering
        \fontsize{9}{11}\selectfont
        \begin{tabular}{ |c|c|c|c|c| } 
            \hline
            & F-mnist & C10 & C100 & Tiny \\
            \hline
            F-mnist & 85.1 $\vert$ {\color{mypurple}79.1} $\vert$ {\color{navy}83.8} & 76.4 $\vert$ {\color{mypurple}67.7} $\vert$ {\color{navy}74.7} & 68.2 $\vert$ {\color{mypurple}65.8} $\vert$ {\color{navy}69.2} & 68.4 $\vert$ {\color{mypurple}61.7} $\vert$ {\color{navy}65.9} \\
            \hline
            C10 & 74.2 $\vert$ {\color{mypurple}71.9} $\vert$ {\color{navy}73.4} & 87.9 $\vert$ {\color{mypurple}84.2} $\vert$ {\color{navy}86.6} & 84.9 $\vert$ {\color{mypurple}82.3} $\vert$ {\color{navy}84.0} & 83.4 $\vert$ {\color{mypurple}80.1} $\vert$ {\color{navy}83.4} \\ 
            \hline
            C100 & 70.8 $\vert$ {\color{mypurple}68.9} $\vert$ {\color{navy}71.2} & 85.0 $\vert$ {\color{mypurple}82.5} $\vert$ {\color{navy}84.3} & 87.0 $\vert$ {\color{mypurple}84.4} $\vert$ {\color{navy}86.5} & 85.6 $\vert$ {\color{mypurple}82.7} $\vert$ {\color{navy}84.8} \\ 
            \hline
            Tiny & 72.3 $\vert$ {\color{mypurple}62.5} $\vert$ {\color{navy}70.9} & 83.6 $\vert$ {\color{mypurple}81.1} $\vert$ {\color{navy}83.6} & 85.3 $\vert$ {\color{mypurple}83.0} $\vert$ {\color{navy}85.0} & 87.9 $\vert$ {\color{mypurple}84.0} $\vert$ {\color{navy}87.3} \\
            \hline
        \end{tabular}
        \caption{Accuracy (\%).}
        \label{subtab:vs_novs_acc}
    \end{subtable}

    \begin{subtable}{\textwidth}
        \centering
        \fontsize{9}{11}\selectfont
        \begin{tabular}{ |c|c|c|c|c| } 
            \hline
            & F-mnist & C10 & C100 & Tiny \\
            \hline
            F-mnist &{1.00} $\vert$ {\color{mypurple}1.00} $\vert$ {\color{navy}1.00} & {0.93} $\vert$ {\color{mypurple}0.90} $\vert$ {\color{navy}0.93} &  {0.94}  $\vert$ {\color{mypurple}0.88}  $\vert$ {\color{navy}0.95} & {0.96} $\vert$ {\color{mypurple}0.90} $\vert$ {\color{navy}0.95}
            \\ \hline
            C10  & {0.80} $\vert$  {\color{mypurple}0.78} $\vert$ {\color{navy}0.82} & {0.99}  $\vert$ {\color{mypurple}0.98} $\vert$ {\color{navy}0.98} & {0.97} $\vert$ {\color{mypurple}0.80} $\vert$ {\color{navy}0.95} & {0.95}  $\vert$ {\color{mypurple}0.90} $\vert$ {\color{navy}0.91}\\  \hline
            C100   & {0.36} $\vert$ {\color{mypurple}0.35} $\vert$ {\color{navy}0.41} & {{0.60}  $\vert$ {\color{mypurple}0.45}} $\vert$ {\color{navy}0.41} & {0.65} $\vert$ {\color{mypurple}0.45} $\vert$ {\color{navy}0.50} & {0.63} $\vert$ {\color{mypurple}0.40} $\vert$ {\color{navy}0.60}\\ \hline
            Tiny  & {0.41} $\vert$ {\color{mypurple}0.10} $\vert$ {\color{navy}0.50} &{{0.61}  $\vert$ {\color{mypurple}0.40}} $\vert$ {\color{navy}0.58} & {{0.57}  $\vert$  {\color{mypurple}0.43}} $\vert$ {\color{navy}0.55} & {0.69} $\vert$ {\color{mypurple}0.50} $\vert$ {\color{navy}0.64} \\
            \hline
        \end{tabular}
        \caption{Average Precision@10. }
        \label{subtab:vs_novs_prec10}
    \end{subtable}
    
    \caption{Performance measures of the predictor with vertex shapes and early stopping (black values), without vertex shapes (purple), without early stopping (blue). The datasets used for training are displayed in the columns, while the rows show the validation ones. Best results are obtained when both vertex shapes and early stopping techniques are employed. }\vspace{-0.5cm}
\end{table*}
\textbf{Network Specialization} We compared through NDCG the ranking induced by the validation accuracy v\_acc${_i}$ with $i=1,\dots,120$ with v\_acc$_{120}$ at the end of training. Fig.~\ref{fig:NDCGat10_2092} highlights a possible correlation between the change of rank and the epoch where the learning rate is dropped (we refer to Appendix~\ref{app:impl_det} for details on hyper-parameters). Moreover, focusing on the NDCG@10, that considers only the top-10, the plot reaches zero way before the end of the training. Therefore, we formulated the hypothesis that the difference between NDCG\@2092 and the NDCG\@10 could be caused by the average performing networks, which are more strongly influenced by the hyper-parameters. We checked the evolution of ranking during training (\ref{fig:evo-rankingCUM},~\ref{fig:evo-rankingDEV}) by counting how many times the architectures change the relevance value (Eq.~\ref{eq:NDCG}). We observe that top-performing
DNNs stop changing rank at earlier epochs, average performing ones have the peak shifted towards the end of training, and DNNs not able to solve the task are such since the early epochs. \vspace{-0.1cm}
\subsection{GRASP-GCN}\label{subsec:grasp_gcn}
We use the validation accuracy, and the Precision@10 measures to evaluate the performance of GRASP-GCN. Tab~\ref{tab:vs_acc}, Tab.~\ref{tab:vs_prec10} show the results obtained for each training dataset (columns) on each of the four validation datasets (rows), while~\cref{fig:early_stopping} gives a compact representation of the GCN performance trained with the accuracy of every training epoch.
The values in the table are above 80 \% when Fashion-MNIST is not involved, which suggests that the predictor trained
over the datasets involved in the lower part of the table, e.g. Cifar-10, Cifar-100, Tiny-
ImageNET, can transfer knowledge over one of these same datasets. On the other hand,
when Fashion-MNIST is involved, either as the training or the validation dataset, the
performance drops drastically compared to the average performance the predictor has with Cifar-10/Cifar-100/Tiny-ImageNet. It is worth noting however, that despite the drop in performance, when Fashion-MNIST is the validation dataset (first row) the drop does not represent a problem. Indeed as the Precision@10 highlights, the top-performing architecture are still correctly ranking. More precisely, as almost ll architectures well-solve Fashion-mnist, a wrong ranking will probably affect architecture with a small difference in validation accuracy, thus resulting in a good ranking order. Finally, Tab.~\ref{tab:vs_sota} shows that GRASP-GCN surpasses all other methods when trained and evaluated over Cifar-10 and when directly applied to new datasets. 

\begin{table}[H]
\centering\fontsize{5.5}{11}\selectfont
\begin{tabular}{ |c|c|c|c|c| } 
\hline
& F-mnist & C10 & C100 & Tiny \\
\hline
F-mnist &{84.1} $\vert$ {\color{myblue}82.2} $\vert$ {\color{red}79.8} &{74.4} $\vert$ {\color{myblue}71.1} $\vert$ {\color{red}68.2} &  {68.2}  $\vert$ {\color{myblue}64.8} $\vert$  {\color{red}68.2} & 68.4 $\vert$ {\color{myblue}63.5} $\vert$ {\color{red}65.5}
 \\ \hline
C10  &{74.2} $\vert$ {\color{myblue} 73.8} $\vert$  {\color{red}73.2} & {87.9}  $\vert$ {\color{myblue}84.1}  $\vert$ {\color{red}86.2} & {84.9} $\vert$ {\color{myblue}81.1} $\vert$ {\color{red}80.8} & {83.4}  $\vert$ {\color{myblue}81.1} $\vert$ {\color{red}80.7} \\  \hline
C100   & {70.8} $\vert$ {\color{myblue}68.5} $\vert$ {\color{red}70.7} & {85.0}  $\vert$ {\color{myblue}80.1} $\vert$ {\color{red}79.1} & {87.0} $\vert$ {\color{myblue}82.1} $\vert$ {\color{red}84.3} & {85.6} $\vert$ {\color{myblue}82.4} $\vert$ {\color{red}85.0} \\ \hline
Tiny  & {72.3} $\vert$ {\color{myblue}66.4} $\vert$ {\color{red}70.1} &{83.6}  $\vert$ {\color{myblue}79.2} $\vert$ {\color{red}83.2} & {85.3}  $\vert$ {\color{myblue} 80.1}  $\vert$  {\color{red}83.4} & {87.9} $\vert$ {\color{myblue}84.0} $\vert$ {\color{red}86.6} \\
\hline
\end{tabular}
\caption{Comparison between (black) ours, (blue) BRP-NAS, (red), MetaD2A.}
\label{tab:vs_sota}\vspace{-0.3cm}
\end{table}

\subsection{Ablation}\label{subsec:ablation}
Tab.~\ref{subtab:vs_novs_acc} highlights how the predictor significantly improves the performance when trained and validated over the same dataset (diagonal values) improving the baseline of more than 3 \%. If we focus on transferability, looking out of the diagonal, we have even larger improvements, with gaps exceeding 9\%. Consistent results are obtained in~\ref{subtab:vs_novs_prec10} with Precision@10 measure.
Finally, Fig.~\ref{fig:early_stopping} ablates on the early-stopping mechanism we propose to employ in the NAS procedure. We can observe that if we use as training set the accuracies the architectures have after the first drop of the learning rate (epoch 40) the performance of the predictor is not affected. This is further validated by Tab.~\ref{subtab:vs_novs_acc}, which compares our best results with and without early stopping.
\vspace{-0.3cm}
\section{Conclusions}
In our work, we face the problem of analyzing the transferability of a predictor under data distribution shift. For this reason, we created our small dataset composed of architectures trained on four datasets. Our ranking analysis on the trained networks showed an association between the drop of the learning rate and the epoch where architectures stop changing their ranking during training, highlighting that top-performing architectures keep their ranking since early training epochs, while the worst ones are such since the early epochs. We moreover spotted that the ranking induced by complex datasets generalizes better than those induced by simple datasets. Given this, we improve the naive predictor training by including the vertex shapes as input, and employing an early stopping procedure. Our method surpasses state-of-the art performance on predictor-based algorithms addressed by distribution shifts. We believe that such a result can help during the search phase of NAS, as to get the true validation accuracy of the top-k ranked architectures to pick the best-performing ones the training can be stopped early reducing significantly the searching time.

\textbf{Limitations and future works} Possible future works could include enlarging the datasets by considering new tasks such as object detection and new modalities such as videos. Another interesting direction could be studying how the method improves with the inclusion of gradients information as input to the predictor.
\section*{Acknowledgement}
This work has been partially supported by the Spanish project PID2022-136436NB-I00, by ICREA under the ICREA Academia programme, and  by unibz startup fund IN 2814 and IN 2902.

{
    \small
    \bibliographystyle{ieeenat_fullname}
    \bibliography{main}
}


\end{document}